\documentclass[
reprint,amsmath,amssymb,aps,pra
]{revtex4-1}

\usepackage{graphicx}
\usepackage{dcolumn}
\usepackage{bm}
\usepackage{graphicx, mathtools}

\usepackage{amsthm, amssymb,enumitem,kantlipsum}

\usepackage{amsthm}

\usepackage{amsfonts}

\usepackage{amssymb}

\usepackage{xcolor}

\usepackage{dsfont}

\newtheorem{theorem}{Theorem}

\usepackage{soul}
\newtheorem{lemma}[theorem]{Lemma}

\usepackage{tabularx}

\usepackage{hyperref}       

\usepackage{url}            

\usepackage{booktabs}       

\usepackage{amsfonts}       

\usepackage{nicefrac}       

\usepackage{microtype}      

\usepackage{xcolor}         

\usepackage{graphicx}

\usepackage{amsmath}

\usepackage{amsthm}

\usepackage{amssymb}

\usepackage{booktabs}

\usepackage{graphicx,subcaption}

\usepackage{enumitem,kantlipsum}

\newcommand{\diag}{\mathop{\mathrm{diag}}}

\newcommand{\trace}{\mathop{\mathrm{tr}}}

\begin{document}
\preprint{APS/123-QED}

\title{Distributionally Robust Domain Adaptation}
\author{Akram S. Awad$^{1}$,   George K. Atia$^{1,2}$}

\affiliation{$^{1}$Department of Electrical and Computer Engineering\\
$^{2}$Department of Computer Science\\
University of Central Florida, Orlando, FL, USA\\  
Emails: akramawad@knights.ucf.edu, george.atia@ucf.edu
}



\begin{abstract}
Domain Adaptation (DA) has recently received significant attention due to its potential to adapt a learning model across source and target domains with mismatched distributions. Since DA methods rely exclusively on the given source and target domain samples, they generally yield models that are vulnerable to noise and unable to adapt to unseen samples from the target domain, which calls for DA methods that guarantee the robustness and generalization of the learned models. In this paper, we propose DRDA, a distributionally robust domain adaptation method. DRDA leverages a distributionally robust optimization (DRO) framework to learn a robust decision function that minimizes the worst-case target domain risk and generalizes to any sample from the target domain by transferring knowledge from a given labeled source domain sample.
 We utilize the Maximum Mean Discrepancy (MMD) metric to construct an ambiguity set of distributions that provably contains the source and target domain distributions with high probability. Hence, the risk is shown to upper bound the out-of-sample target domain loss.   
 Our experimental results demonstrate that our formulation outperforms existing robust learning approaches. 
 \end{abstract}

\maketitle

\onecolumngrid

\section{Introduction}

    The performance of conventionally trained machine learning models can significantly degrade when the distribution of the data at the time of inference is different from that at the time of training. 
    Domain adaptation (DA) is concerned with adapting learning algorithms trained in a source domain using samples from a given distribution to a target domain where the test samples are drawn from a different distribution \citep{weiss2016survey}. These distributions are referred to as the source and target domain distribution, respectively. 
    Given its ability to mitigate the distributional mismatch, DA has made significant strides in diverse application domains, including but not limited to computer vision \citep{pereira2018semi,6126286,tzeng2017adversarial,fernando2013unsupervised}, natural language processing \citep{mou2016transferable,guo2020multi,han2019unsupervised}, and regression analysis  \citep{cortes2011domain,cortes2014domain}.
    
    The key challenge underlying DA is to reduce the discrepancy between the source and target domain distributions, which has been tackled using a number of approaches. One main approach is \textit{instance weighting} in which the source sample instances are re-weighted to minimize the distribution mismatch while learning a decision function  \cite{sugiyama2007direct,huang2006correcting,bruzzone2009domain}. An alternative strategy is to find across-domain feature representations that simultaneously minimize the discrepancy between distributions and preserve intrinsic statistical 
    and structural properties of the data \citep{long2013transfer,zhang2017joint,long2014domain}. The main shortcoming of the foregoing approaches is that the decision function they learn is often insufficiently robust to generalize to unseen samples from the target domain. This is largely because they minimize the discrepancy between the empirical distributions associated with the given source and target samples rather than the true population distributions.
    In turn, the learned decision function 
    has propensity for unpredictable performance 
    in the presence of noise or with out-of-sample data. This could have drastic impact on AI systems for autonomous driving, automation, and surveillance which, in addition to knowledge transfer across disparate (but related) domains, prioritize the safety and robustness to perturbations that disrupt normal operation. This motivates our work on developing robust versions of DA methods with out-of-sample performance guarantees. 
    \par 
    Distributionally Robust Optimization (DRO) is the problem of finding the optimal decision function that minimizes the worst-case risk over an uncertainty (or ambiguity) set of distributions. The DRO framework has 
    gained attention in the context of 
    robust optimization 
    \citep{ben2009robust,bertsimas2004price} and more recently adversarial learning \citep{staib2017distributionally,madry2017towards}.    
    Several ways have been proposed in the literature to construct such ambiguity sets. One approach uses \textit{moment-based} ambiguity sets, which include all distributions that satisfy certain statistical properties in the form of some moment constraints \citep{delage2010distributionally,goh2010distributionally}. An alternative approach -- the focus of this work -- constructs \textit{distance-based} ambiguity sets, which define a ball of distributions that are within a certain distance with respect to some discrepancy metric from an empirical distribution. A key result of the latter is that, if the ambiguity set is large enough to contain the true population distribution with high probability, then the DRO objective (i.e., the worst-case risk) gives a high-probability upper bound on the population risk.  

    Different discrepancy metrics have been used to construct the ambiguity set such as the Wasserstein distance \citep{pflug2007ambiguity,gao2016distributionally} and the Kullback–Leibler divergence \citep{ben2013robust,lam2016robust}. While this choice of metrics is motivated by a number of structural results that facilitate the solution to the DRO formulation, the resulting ambiguity sets have main drawbacks. 
    The Kullback–Leibler divergence set contains only discrete distributions with the same finite support as the empirical distribution, which makes it unsuitable when the true population distribution is continuous. 
    The Wasserstein ambiguity set is computationally expensive and, more importantly, its radius has to scale with the data dimension to certify out-of-sample performance. To address these limitations, \citep{staib2019distributionally} defines the ambiguity set with respect to the Maximum Mean Discrepancy (MMD) \citep{gretton2012kernel}, resulting in an optimization over embedding means of distributions. 
    The MMD DRO averts the aforementioned drawbacks since the MMD ambiguity set contains both discrete and continuous distributions and its radius is independent of the data dimension. The unifying work of \citep{zhu2020kernel} introduces a wide range of kernel-based ambiguity sets and relaxes the assumptions on the loss function in the DRO formulation. \par

     Here, we propose a Distributionally Robust Domain Adaptation (DRDA) framework. The main objective of our formulation is to learn a robust and generalized regression function that generalizes well on a target domain given a labeled and an unlabeled sample from the source and target domains, respectively. Since the target domain data is unlabeled, DRDA adapts the knowledge of the source domain sample by re-weighting its instances and simultaneously learning a robust decision function.
    Some previous works have considered the use of DRO across domains for DA \citep{liu2014robust,chen2016robust,taskesen2021sequential,liu2017kernel,wang2020distributionally}. There, the  search is over an uncertainty set of probabilistic (conditional) mappings from input to output subject to moment constraints. 
    In sharp contrast, here we define an uncertainty set of joint distributions within a given distance from a weighted empirical source domain distribution with respect to the MMD metric, with the main goal of establishing guarantees on out-of-sample performance.

     The conventional DA framework does not account for uncertainty in the given source and target domain samples (e.g., due to contamination with noise), wherefore the model it learns could yield inaccurate predictions. DRO is not directly applicable in the target domain since the labeling information is unavailable. On the other hand, DRDA addresses these limitations by transferring knowledge within uncertainty sets of distributions from the labeled source domain, thereby simultaneously accounting for the discrepancy between domains and the uncertainty in the target domain.
     The contributions of our work can be summarized as follows.
    \begin{enumerate}
        \item We formulate a robust DA problem, dubbed Distributionally Robust Domain Adaptation (DRDA), to learn a robust regression model that guarantees the out-of-sample performance in the target domain.
        \item We construct an MMD ambiguity set and prove that it contains the source and target domain distributions with high probability, thereby ensuring the generalization of the learned model. 
        \item We develop a solution methodology to the formulated DRDA problem   
        leveraging a DRO formulation under an additional common assumption on the loss function. 
       
    \end{enumerate}

\section{Background}
\subsection{Notation}
Let $\mathcal{X}\subseteq\mathbb{R}^N$ be a topological input space, and $\mathcal{P}$ the set of all probability measures defined on $\mathcal{X}$. Let $l_{h}$ be a loss function associated with a decision function $h$, henceforth, we denote it by $l_h$ for simplicity, and $d :\mathcal{P}\times\mathcal{P}\to \mathbb{R}$ a distance metric between probability measures. We denote the pair $(x,y)$ by $\zeta$. We use $\hat{\mathbf{P}}(.)= \frac{1}{m}\sum_{j=1}^{m}\delta_{\zeta_{j}}(.)$ to represent the empirical distribution on the sample $\{\zeta_j\}_{j=1}^{m}$, where $\delta$ is the Dirac measure.
Let $\mathcal{H}$ be a Reproducing Kernel Hilbert Space (RKHS) associated with a characteristic kernel $k$, $\phi:\mathbb{R}^N\to \mathcal{H}$ is the corresponding feature map, $\langle.,.\rangle_\mathcal{H}$ is the inner product on $\mathcal{H}$ and $\|.\|_\mathcal{H}$ is the induced norm.
We define $\mu_\mathbf{P}=\mathbb{E}^ \mathbf{P}[\phi(x)]$ to be the embedding mean of the probability measure $\mathbf{P}$, where $\mu_\mathbf{P} \in \mathcal{H}$ and $\mathbb{E}^ \mathbf{P}[.]$ denotes the expectation with respect to measure $\mathbf{P}$. Since the kernel $k$ is assumed to be characteristic, every probability measure $\mathbf{P}\in\mathcal{P}$ is embedded as a unique element in $\mathcal{H}$ \citep{muandet2017kernel}. Hence,
the embedding mean $\mu_\mathbf{P}$ is an injective map. 

\subsection{Domain Adaptation (DA)}
In the DA setting, we are given a labeled source domain sample $D_s=\{x_i,y_i\}_{i=1}^{n_s}$ and an unlabeled target domain sample $D_t=\{x_i\}_{i=1}^{n_t}$ drawn from two different distributions, where $n_s$ and $n_t$ are the source and target domain sample sizes, respectively.  
 In DA, one seeks to find a decision function $h$ that minimizes the target domain risk $R_t=\mathbb{E}^ {\mathbf{P_t}}[l_{h}]$. 
 Since the labeling information is unavailable for the target domain sample, DA transfers knowledge from the source domain to improve the performance of the learned decision function in the target domain.   
    
\smallbreak
\noindent\textbf{Covariate shift and density ratio:}
Under the covariate shift assumption, it is assumed that
the conditional distributions of the labels given the features are similar across domains, but the marginals are different \cite{huang2006correcting}. 
We define the density ratio between the target and source domain distributions 
as $w(\zeta)= \frac{\mathbf{P_t}(\zeta)}{\mathbf{P_s}(\zeta)}$,
where $\mathbf{P_t}$ is absolutely continuous with respect to $\mathbf{P_s}$. 

Let $\hat{w}(\zeta)$ be an estimate of the density ratio $w(\zeta)$, so we define $\mathbf{P_{t'}}(\zeta)= \hat{w}(\zeta)\mathbf{P_s}(\zeta) $ as the weighted source domain probability distribution and  $\hat{\mathbf{P} }_{t'}(.)= \frac{1}{n_s} \sum_{i=1}^{n_s}\hat{w}(\zeta_i)\delta_{\zeta_i}(.)$ as its corresponding empirical distribution. 

The Maximum Mean Discrepancy (MMD) distance, denoted by $d_m$, between two probability distributions $\mathbf{Q}$ and $\mathbf{P}$ is defined as
\begin{equation}
\begin{gathered}
    d_m(\mathbf{Q},\mathbf{P}) = \sup_{\|f\|_{\mathcal{H}}\leq1} \mathbb{E}^\mathbf{Q}[f(x)]-\mathbb{E}^\mathbf{P}[f(x)] \\
        = \sup_{\|f\|_{\mathcal{H}}\leq1} <f, \mu_\mathbf{Q} -\mu_\mathbf{P}>_\mathcal{H} 
        = \|\mu_\mathbf{Q} -\mu_\mathbf{P}\|_\mathcal{H} \:. 
\end{gathered}
\end{equation}

Since 
the embedding mean is an injective map when the kernel associated with the RKHS is characteristic, the MMD metric can measure the distance between distributions by finding the distance between their embedding means.

\smallbreak

\noindent\textbf{Kernel Mean Matching (KMM)} \cite{huang2006correcting} is concerned with finding the weights $\hat{w}(x)\leq B$ such that the MMD distance between the weighted source and target probability measures is minimized. Thus, the KMM problem is defined as follows
 \begin{equation} \label{eqn: kmm}
 \begin{gathered}
          \min_{\hat{w}(x)} \|\mu_\mathbf{P_{t'}} -\mu_\mathbf{P_t}\|_\mathcal{H}
          \quad\text{subject to} \quad \hat{w}(x)\in[0,B],\;\\
          \int \hat{w}(x) d\mathbf{P_{s}}=1\:.
 \end{gathered}
 \end{equation}

\subsection{Distributionally Robust Optimization (DRO)}
Let $\zeta_1,\dots, \zeta_n$ be an i.i.d. sample drawn from a probability distribution $\mathbf{P}\in\mathcal{P}$. The Distributionally Robust Optimization (DRO) problem is defined as
\begin{equation} \label{eqn: General DRO}
    \hat{J_n}= \inf_{h} \sup_{\mathbf{Q}\in \Omega} \mathbb{E}^ \mathbf{Q}[l_{h}]\:,
\end{equation}
 where $\Omega$ is the ambiguity set 
\begin{equation}
  \Omega = \{\mathbf{Q} | d(\mathbf{Q},  \hat{\mathbf{P}}) \leq \epsilon \} \:.
\end{equation}
 The DRO problem \eqref{eqn: General DRO} finds the learning model $h$ that minimizes the worst-case risk. 
 Specifically, it optimizes over all distributions in the ambiguity set, i.e., that are within a distance $\epsilon$ from the empirical distribution $\hat{\mathbf{P}}$. A key challenge in the DRO problem is to construct an ambiguity set that contains the true population distribution with high confidence. Formally, if we ensure that $\mathbf{P}\in\Omega$ with high probability, then fixing any model $h$, $ \mathbb{E}^ \mathbf{P}[l_{h}]\leq\sup_{\mathbf{Q}\in \Omega} \mathbb{E}^ \mathbf{Q}[l_{h}]$ with high probability. Therefore, if the ambiguity set is chosen appropriately, the DRO problem gives a high probability bound on the true population risk.
 
\section{Distributionally Robust Domain Adaptation (DRDA)}
In this section, we present the problem setup, formulate the DRDA problem, and establish our main theoretical results.

We are given two samples $D_s$ and $D_t$ in the source and target domains, respectively, where the labels of $D_t$ are not available. The samples are drawn from probability measures $\mathbf{P_s}$ and $\mathbf{P_t}$, respectively. The goal of DRDA is to learn a hypothesis $h : \mathcal{X}\to\mathbb{R}$ that simultaneously minimizes the target domain risk $R_t=\mathbb{E}^ {\mathbf{P_t}}[l_{h}]$, and generalizes well on any unseen sample from the target domain distribution $\mathbf{P_t}$.

Towards this goal, 
we seek to solve the following DRO problem
\begin{equation} \label{eqn: Tgt DRO}
    \hat{J}_{n_t}= \inf_{h} \sup_{\mathbf{Q}\in \Omega_t} \mathbb{E}^ \mathbf{Q}[l_{h}],
\end{equation}
where the ambiguity set $\Omega_t :=\{\mathbf{Q} | d(\mathbf{Q},  \hat{ \mathbf{P}}_t) \leq \epsilon_t \}$ is centered at the empirical target domain probability measure $\hat{\mathbf{P} }_t(.)= \frac{1}{n_t} \sum_{i=1}^{n_t}\delta_{\zeta_i}(.)$. The DRO formulation in \eqref{eqn: Tgt DRO}  finds the decision function $h$ that minimizes the worst-case target domain risk. In principle, if we ensure that $\mathbf{P_t}\in \Omega_t$ with high confidence, then $\mathbb{E}^ {\mathbf{P_t}}[l_{h}]\leq\sup_{\mathbf{Q}\in \Omega_t} \mathbb{E}^ \mathbf{Q}[l_{h}]$ with high probability. 

However, a key difficulty for establishing such guarantee is that the labels for the given target domain sample $D_t$ are not available. Therefore, we make use of the labeled source domain sample $D_s$ to learn a hypothesis $h$ that achieves the desired two-fold objective. Hence, we propose the following DRO problem
\begin{equation} \label{eqn: W DRO}
    \hat{J}_{n_{t'}}= \inf_{h} \sup_{\mathbf{Q}\in \Omega_{t'}} \mathbb{E}^ \mathbf{Q}[l_{h}]
\end{equation}
where $\Omega_{t'} =\{\mathbf{Q} | d(\mathbf{Q},  \hat{\mathbf{P} }_{t'}) \leq \epsilon_{t'} \}$, recalling that $\hat{\mathbf{P} }_{t'}$ is the empirical weighted source domain probability. The introduced set $\Omega_{t'}$ can be viewed as a transferred version of a source domain ambiguity set $\Omega_{s}$ (that is centered at the empirical source domain distribution $\hat{\mathbf{P} }_{s}$). However, how do we set the radius $\epsilon_{t'}$ to ensure that $\mathbf{P_t}, \mathbf{P_{t'}}\in \Omega_{t'}$ with high confidence? 
We answer this question by establishing the following results.


\begin{lemma} \label{L1}
Let $k: \mathcal{X}\times\mathcal{X}\to \mathbb{R}$ be a positive definite kernel on the space $\mathcal{X}$ with $\|\phi(x)\|_\mathcal{H}\leq \sqrt{M}, \forall x \in \mathcal{X}$. Let  $\hat{w}(x)\in [0,B]$, and $0<\delta<1$. Then, with probability at least $1-\delta$, 
\begin{equation}\label{eqn: B0}
    d_m(\mathbf{P_{t'}},\hat{\mathbf{P}}_{t'}\ )\leq B\sqrt{\frac{M}{n_s}} \left(1+\sqrt{2\log\frac{1}{\delta}} \right).
\end{equation}
\proof
We make use of Mcdiarmid’s inequality as in the proof of the concentration result \cite[Proposition A.1]{tolstikhin2017minimax}. Let $x_1,\dots,x_n$ be independent random variables, and let $f: \mathcal{X}_1\times\dots\times\mathcal{X}_n\to \mathbb{R}$ be a function, such that for every $(x_1,\dots,x_n), (x'_1,\dots,x'_n)\in\mathcal{X}_1\times\dots\times\mathcal{X}_n $ that differ only in the $i^{th}$ element ($x_l=x'_l, \forall l\neq i$),  we have
\begin{equation*}
    |f(x_1,\dots,x_n)-f(x'_1,\dots,x'_n)|\leq d_i\:.
\end{equation*} 
Therefore, for any $t\geq0$, 
\begin{align}\label{eqn: Mcdiarmid}
    \Pr[f(x_1,\dots,x_n)-\mathbb{E} f(x_1,\dots,x_n)\geq t]\leq \exp{\bigg(\frac{-2t^2}{\sum\limits_{i=1}^{n} d^2_i}\bigg)}.
\end{align}
Let 
\begin{equation*}
\begin{gathered}
     f(x_1,\dots,x_n)= d_m(\mathbf{P_{t'}},\hat{\mathbf{P}}_{t'}\ )= \|\mu_{\mathbf{P_{t'}}}-\mu_{\hat{\mathbf{P}}_{t'}}\|_\mathcal{H}\\
     f(x'_1,\dots,x'_n)= d_m(\mathbf{P_{t'}},\hat{\mathbf{P'}}_{t'}\ )=\|\mu_{\mathbf{P_{t'}}}-\mu_{\hat{\mathbf{P'}}_{t'}}\|_\mathcal{H}
\end{gathered}
\end{equation*}
where $\hat{\mathbf{P'}}_{t'}=\frac{1}{n_s}\sum\limits_{i=1}^{n_s}\hat{w}(x'_i)\delta(x'_i)$.\\

First, we find the value of $d_i$. 
We have that 
\begin{equation*}
\begin{gathered}
     |\|\mu_{\mathbf{P_{t'}}}-\mu_{\hat{\mathbf{P}}_{t'}}\|_\mathcal{H}-\|\mu_{\mathbf{P_{t'}}}-\mu_{\hat{\mathbf{P'}}_{t'}}\|_\mathcal{H}|\leq  |\sup_{\|f\|_{\mathcal{H}}\leq1} \mathbb{E}^\mathbf{\hat{\mathbf{P}}_{t'}}f(x)-\mathbb{E}^\mathbf{\hat{\mathbf{P'}}_{t'}}f(x) |
     = \|\mu_{\hat{\mathbf{P}}_{t'}}-\mu_{\hat{\mathbf{P'}}_{t'}}\|_\mathcal{H}\\
     = \frac{1}{n_s}\|\hat{w}(x_i)\phi(x_i)-\hat{w}(x'_i)\phi(x'_i)\|_\mathcal{H}\leq \frac{2B\sqrt{M}}{n_s}\\
\end{gathered}
\end{equation*}
By setting the RHS of \eqref{eqn: Mcdiarmid} to $\delta\in (0,1)$, it holds with probability at least $1-\delta$ that
\begin{equation*}
    \|\mu_{\mathbf{P_{t'}}}-\mu_{\hat{\mathbf{P}}_{t'}}\|_\mathcal{H}-\mathbb{E}\|\mu_{\mathbf{P_{t'}}}-\mu_{\hat{\mathbf{P}}_{t'}}\|_\mathcal{H}\leq B\sqrt{\frac{2M}{n_s}\log\frac{1}{\delta}}.
\end{equation*}

Now,
\begin{equation*}
\begin{gathered}
     \mathbb{E}\|\mu_{\mathbf{P_{t'}}}-\mu_{\hat{\mathbf{P}}_{t'}}\|_\mathcal{H}\leq\sqrt{\mathbb{E}\|\mu_{\mathbf{P_{t'}}}-\mu_{\hat{\mathbf{P}}_{t'}}\|^2_\mathcal{H}}  
  {\buildrel (a)\over\leq} \sqrt{\frac{\mathbb{E}_{x\sim\mathbf{P}_{s}}\hat{w}^2(x)\|\phi(x)\|^2_{\mathcal{H}}\hspace{-0.75mm}-\hspace{-0.75mm}\|\hspace{-0.75mm}\int_\mathcal{X}\hspace{-.5mm}\hat{w}(x)\phi(x)d\mathbf{P}_{s}\|^2_{\mathcal{H}}}{n_s}}\leq \hspace{-.75mm}B\hspace{-.25mm} \sqrt{\frac{M}{n_{s}}} 
\end{gathered}
\end{equation*}
Inequality (a) follows the same reasoning in \cite{tolstikhin2017minimax}. 
\qed
\end{lemma}

Lemma \ref{L1} establishes a high probability upper bound on the distance between the weighted source domain probability and its empirical version, thus can guide the choice of the radius of $\Omega_{t'}$ so that $\mathbf{P}_{t'}\in \Omega_{t'}$ with high probability.
However, this does not guarantee that the target domain population distribution $\mathbf{P}_{t}\in \Omega_{t'}$ with high probability. 
%

%
Next, we present our main result for setting the radius $\epsilon_{t'}$ of the proposed ambiguity set $\Omega_{t'}$. The choice of $\epsilon_{t'}$ in the statement of Theorem \ref{Thm} guarantees the generalization of the learned model with high probability.   

\begin{theorem}\label{Thm}
Let $w\in [0,B]$, then for any $\delta\in(0,1)$, set the radius $\epsilon_{t'}$ of $\Omega_{t'}$ in \eqref{eqn: W DRO} as
\begin{align}\label{eqn: main results}
     \epsilon_{t'}=\sqrt{M}\left(\sqrt{\frac{B^2}{n_s}+\frac{1}{n_t}}+\sqrt{\frac{1}{n_t}}\right ) \left(1+\sqrt{2\log\frac{1}{\delta}} \right).
\end{align}
Then, 
   with probability at least $1-\delta$, $\mathbf{P_t}\in \Omega_{t'}$  and $\mathbb{E}^ {\mathbf{P_{t}}}[l_{h}]\leq\sup_{\mathbf{Q}\in \Omega_{t'}} \mathbb{E}^ \mathbf{Q}[l_{h}]$.\\  \label{s1} 
    

\proof
Let's define $\hat{\mathbf{P} }_{t'}(.)= \frac{1}{n_s} \sum_{i=1}^{n_s}w (\zeta_i)\delta_{\zeta_i}(.)$.  First, we obtain a high probability upper bound on the distance between the empirical target and the weighted source probability measures. For any $\delta\in (0,1)$, 
\begin{equation}\label{eqn: B1}
    d_m(\hat{\mathbf{P}}_{t},\hat{\mathbf{P}}_{t'} )\leq \sqrt{M}\sqrt{\frac{B^2}{n_s}+\frac{1}{n_t}} \left(1+\sqrt{2\log\frac{1}{\delta}} \right)
\end{equation}
with probability at least $1-\delta$, from the result of \cite{huang2006correcting}, which characterizes the deviation between empirical means across domains. By invoking the concentration of measure result in \cite{tolstikhin2017minimax}, we get that 
\begin{equation} \label{eqn: Epst}
         d_m(\mathbf{P_{t}},\hat{\mathbf{P}}_{t}\ )\leq \sqrt{\frac{M}{n_t}} \left(1+\sqrt{2\log\frac{1}{\delta}} \right).
\end{equation}
From the triangular inequality,
\begin{align}\label{eqn: triangluar}
    d_m(\mathbf{P_{t}},\hat{\mathbf{P}}_{t'} ) & \leq  d_m(\hat{\mathbf{P}}_{t},\hat{\mathbf{P}}_{t'} ) + d_m(\mathbf{P_{t}},\hat{\mathbf{P}}_{t} )
\end{align}

 Therefore, by summing \eqref{eqn: B1} and \eqref{eqn: Epst}, we obtain an upper bound on $d_m(\mathbf{P_{t}},\hat{\mathbf{P}}_{t'} )$. Hence, we set $\epsilon_{t'}$ equal to this upper bound as in \eqref{eqn: main results}. The result in \ref{s1} follows since $\epsilon_{t'}$ bounds $d_m(\mathbf{P_{t}},\hat{\mathbf{P}}_{t'} )$ from above with probability $1-\delta$, i.e., 
 $\mathbf{P_t}\in \Omega_{t'}$ and the target domain risk is less than the worst-case risk on $\Omega_{t'}$ with probability $1-\delta$. 
 
 
\qed

\end{theorem}

Given the primal robust domain adaptation problem in \eqref{eqn: W DRO}, the value of $\hat{J}_{n_t}$ depends on the density ratio $\hat{w}(x)$,  since we are optimizing over all distributions that lie in the ambiguity set $\Omega_{t'}$.  The main problem in  \eqref{eqn: W DRO} aims at finding the decision function $h$ corresponding to the worst-case distribution such that the discrepancy between the source and target domain probability measures is minimized.  This objective can be achieved in two different ways: 
(i) $\hat{w}$ can be estimated by solving the KMM \eqref{eqn: kmm} and then \eqref{eqn: W DRO} can be used to estimate the decision function $h$; (ii) we can optimize jointly over the decision function $h$ and the density ratio $\hat{w}$, and since 
$\sup_{\mathbf{Q}\in \Omega_{t'}} \mathbb{E}^ \mathbf{Q}[l_{h}]\leq\sup_{\mathbf{Q}\in \Omega_{t'}} \mathbb{E}^ \mathbf{Q}[l_{h}]+ d_m^2 (\mathbf{P}_{t'},\mathbf{P}_{t})$, we can instead solve
\begin{equation}\label{eqn: DRDA}
    \inf_{h,\hat{w}} \sup_{\mathbf{Q}\in \Omega_{t'}} \mathbb{E}^ \mathbf{Q}[l_{h}]+ \beta d_m^2 ({\mathbf{P}}_{t'},{\mathbf{P}}_{t})\:,
\end{equation}
where $\beta\geq0$ is introduced to control the domain adaptation component.\par
Optimizing jointly over $h$ and $\hat{w}$ is desirable since the parameters that control the learned density ratio and the decision function are not independent \cite{bickel2009discriminative}. More specifically, the source domain sample is common in both the estimation of the weights $\hat{w}$ and for learning the decision function $h$. Hence, optimizing first over $\hat{w}$ (using KMM) then over $h$ using the DRO formulation in \eqref{eqn: W DRO} would yield a sub-optimal solution. We refer to the formulation in \eqref{eqn: DRDA} as Distributionally Robust Domain Adaptation (DRDA). The first term in the DRDA formulation accounts for the generalization of the learned model on the target domain, while the second regularizing term is to minimize the discrepancy between the two domain distributions. 
\section{Proposed DRDA Formulation }
In this section, we present a DRO-based formulation to solve the DRDA problem in \eqref{eqn: DRDA} 
%
under the assumption that $l\in \mathcal{H}$. A similar assumption was made in \cite{staib2019distributionally} for the DRO problem.
Since $\mu_\mathbf{P}$ is an injective map, 
we have that 
\begin{align}\label{eqn: eq1}
    \sup_{\mathbf{Q}\in \Omega_{t'}} \mathbb{E}^ \mathbf{Q}[l_{h}]&\leq \sup_{\mu_\mathbf{Q}:\|\mu_\mathbf{Q} -\mu_\mathbf{\hat{\mathbf{P} }_{t'}}\|_{\mathcal{H}}\leq \epsilon_{t'}}\langle\mu_\mathbf{Q},l\rangle_{\mathcal{H}}= \quad\mathbb{E}^ {\hat{\mathbf{P} }_{t'}}[l_{h}]+\epsilon_{t'}\|l\|_\mathcal{H}
\end{align}
The inequality in \eqref{eqn: eq1} is because not every element of the RKHS is an embedding mean of some probability measure. Therefore, we can rewrite the DRDA problem as
\begin{equation*}
    \begin{gathered}
      \inf_{h,\hat{w}(x)} \sup_{\mathbf{Q} \in \Omega_{t'}} \mathbb{E}^ \mathbf{Q}[l_{h}]+  \beta d_m^2 ({\mathbf{P} }_{t'},{\mathbf{P}}_{t})  
   \leq \inf_{h,\hat{w}(x)}\mathbb{E}^{\hat{\mathbf{P} }_{t'}}[l_{h}]+\epsilon_{t'}\|l\|_\mathcal{H}+ \beta d_m^2 (\mathbf{P} _{t'},\mathbf{P}_{t})\:.\\
    \end{gathered}
\end{equation*}
    Since we do not have access to the true underlying source and target domain distributions and only have samples from both domains, we make use of the empirical MMD distance $d_m (\hat{\mathbf{P}}_{t'},\hat{\mathbf{P}}_{t})$. Thus, instead we solve \begin{equation}
    \begin{gathered}
    \inf_{h,\hat{w}(x_i)}\mathbb{E}^{\hat{\mathbf{P}}_{t'}}[l_{h}]+\epsilon_{t'}\|l\|_\mathcal{H}+ \beta d_m^2 (\hat{\mathbf{P} }_{t'},\hat{\mathbf{P}}_{t})\\
     =\inf_{h,\hat{w}(x)} \frac{1}{n_s} \sum\limits_{i=1}^{n_s}\hat{w}(x_i)l_{h}(x_i) +\epsilon_{t'}\|l\|_\mathcal{H} + \quad\beta \left\|\frac{1}{n_s}\sum\limits_{i=1}^{n_{s}}\hat{w}(x_i)\phi(x_i)-\frac{1}{n_t}\sum\limits_{i=1}^{n_{t}}\phi(x_i)\right\|^2_\mathcal{H}\\
    \end{gathered}
\end{equation}

The weights $\hat{w}(x_i)$ are bounded by a constant $B$ per the assumption in Lemma \ref{L1}. Also, the weighted source domain probability must sum up to 1, i.e., $\int_\mathcal{X}\hat{w}(x) d\mathbf{P_s} =1$, thus for the empirical one $\sum\limits_{i=1}^{n_{s}} \hat{w}(x_i)=n_s$. We can readily formulate the final DRDA problem as
\begin{equation} \label{eqn: form1}
    \begin{gathered}
      \inf_{h,\hat{w}(x_i)} \frac{1}{n_s} \sum\limits_{i=1}^{n_s}\hat{w}(x_i)l_h(x_i) +\epsilon_{t'}\|l_h\|_\mathcal{H}
    + \quad \beta \left\|\frac{1}{n_s}\sum\limits_{i=1}^{n_{s}}\hat{w}(x_i)\phi(x_i)-\frac{1}{n_t}\sum\limits_{i=1}^{n_{t}}\phi(x_i)\right\|^2_\mathcal{H} \\ \quad\text{s.t.}
     \quad  \hat{w}(x_i)\in [0,B]\\
     \left|\sum\limits_{i=1}^{n_{s}} \hat{w}(x_i)-n_s\right|\leq n_s c , \quad c>0
    \end{gathered}
\end{equation}
 
\section{Proposed Solution}
In this section, we present our solution to the DRDA formulation in \eqref{eqn: form1}.
We consider the RKHS $\mathcal{H}_\sigma$ induced by a Gaussian kernel $k_\sigma(x,y)=\exp{\bigg(-\frac{\|x-y\|^2}{2\sigma^2}\bigg)}$ of bandwidth $\sigma$, and assume a quadratic loss function $l_{h}=(h(x)-g(x))^2$, where $g(x)$ is a labeling function. We use $\|.\|_\sigma$ and $\langle.,.\rangle_\sigma$ to denote the norm and inner product of the corresponding RKHS $\mathcal{H}_\sigma$, respectively. Therefore, we need to minimize the objective function
\begin{equation}\label{eqn: Reg1}
    \mathbb{E}^ {\hat{\mathbf{P} }_{t'}}[(h(x)-g(x))^2]+\epsilon_{t'}\|(h-g)^2\|_\sigma+\beta d_m^2 (\hat{\mathbf{P} }_{t'},\hat{\mathbf{P} }_{t})\:. 
\end{equation}
To bound the norm $\|(h-g)^2\|_\sigma$, we need the following lemma from \cite[Theorem 4.1]{staib2019distributionally}.
\begin{lemma} \label{L3}
If $h,g\in \mathcal{H}_\sigma$, that is, the RKHS corresponds to the Gaussian kernel, then $ \| h g\|_{\frac{\sigma}{\sqrt{2}} } \leq \| h\|_\sigma \| g\|_\sigma$.
\end{lemma}
Since $fg\in \mathcal{H}_\frac{\sigma}{\sqrt{2}}$, from the triangular inequality, it follows that $\|(h-g)^2\|_{\sigma}\leq\|h^2\|_\frac{\sigma}{\sqrt{2}}+\|g^2\|_\frac{\sigma}{\sqrt{2}}+ 2 \|h\|_\sigma\|g\|_\sigma$. Therefore, the objective function in \eqref{eqn: Reg1} can be bounded by
\begin{equation}\label{eqn: Reg2}
\begin{gathered}
     \mathbb{E}^{\hat{\mathbf{P} }_{t'}}[(h(x)-g(x))^2]+\epsilon_{t'}(\|h^2\|_\frac{\sigma}{\sqrt{2}}+\|g^2\|_\frac{\sigma}{\sqrt{2}}+ 2 \|h\|_\sigma\|g\|_\sigma)+\beta d_m (\hat{\mathbf{P} }_{t'},\hat{\mathbf{P} }_{t})^2
\end{gathered}
\end{equation}

In addition, $d_m (\hat{\mathbf{P} }_{t'},\hat{\mathbf{P}}_{t})$ can be written as
\begin{align}
\begin{gathered}
    d_m (\hat{\mathbf{P} }_{t'},\hat{\mathbf{P}}_{t})=\left\|\frac{1}{n_s}\sum\limits_{i=1}^{n_{s}}\hat{w}(x_i)\phi(x_i)-\frac{1}{n_t}\sum\limits_{i=1}^{n_{t}}\phi(x_i)\right\|^2_\mathcal{H}\\
    = \frac{1}{n^2_s}\hat{w}\mathbf{K}^s_{\sigma}\hat{w}-\frac{2}{n_t n_s}\hat{w}^T\mathbf{K}^{s,t}_\sigma\bold{1}_{n_t}+ \text{const}\:, 
\end{gathered}
\end{align}
and the first term in \eqref{eqn: Reg2} as
\begin{equation}\label{eqn: 23}
\begin{gathered}
     \mathbb{E}^{\hat{\mathbf{P} }_{t'}}[(h(x)-y)^2]= \frac{1}{n_{s}}\sum\limits_{i=1}^{n_s} \hat{w}(x_i) (h(x_i)-y_i)^2
     =\frac{1}{n_{s}} (\mathbf{K}^s_\sigma \alpha-y_s)^T \mathbf{W} (\mathbf{K}^s_\sigma \alpha-y_s)\:,
\end{gathered}
\end{equation}
where the matrix $\mathbf{K}^s_\sigma$ has the elements $k_\sigma(x_i,x_j), i,j=1,\dots, n_s$, the matrix $\mathbf{K}^{s,t}_\sigma$ has the elements $k_\sigma(x_i,x_j),  i=1,\dots, n_s, j=1,\dots, n_t$, $\mathbf{W}=\diag(\hat{w}(x_1), \dots, \hat{w}(x_{n_s}))$ and $y= (y_1,\dots, y_{n_s})$, where $\diag(.)$ returns a diagonal matrix of its vector argument.
Since $h\in \mathcal{H}_{\sigma}$, then by the representer Theorem \cite{Scholkopf01ageneralized}, we have the expansion $h=\sum_{i=1}^{n_s} \alpha_i\phi(x_i)$. Using the bound in \eqref{eqn: Reg2}, the DRDA problem in \eqref{eqn: form1} can be expressed as 
\begin{equation} \label{eqn: DRDA1}
    \begin{gathered}
        \inf_{\alpha, \hat{w}} (\mathbf{K}^s_\sigma \alpha-y_s)^T \mathbf{W} (\mathbf{K}^s_\sigma \alpha-y_s)+  \beta(\frac{1}{n^2_s}\hat{w}\mathbf{K}^s_{\sigma}\hat{w}-\frac{2}{n_t n_s}\hat{w}^T\mathbf{K}^{s,t}_\sigma\bold{1}_{n_t}) + \lambda \trace((\mathbf{D}_\alpha \mathbf{K}^s_\frac{\sigma}{\sqrt{2}})^4) +\lambda \alpha^T \mathbf{K}^s_\sigma \alpha\\
         \text{s.t.}\quad  \hat{w}(x_i)\in [0,B]\\
     \left|\sum\limits_{i=1}^{n_{s}} \hat{w}(x_i)-n_s\right|\leq n_s c , \quad c>0
    \end{gathered}
\end{equation}

where $\alpha=(\alpha_1,\dots,\alpha_{n_s} )^T$, $\mathbf{D}_\alpha= \diag(\alpha_1,\dots,\alpha_{n_s})$. 
\section{Generalization Bound}
In this section, we derive a generalization bound on the true (population) target domain risk  $R_t=\mathbb{E}^ {\mathbf{P_t}}[l_{h}]=\mathbb{E}^ {\mathbf{P_t}}[(h(x)-g(x))^2]$ in terms of the empirical source domain risk $\hat{R}_s=\mathbb{E}^ {\mathbf{\hat{P}_s}}[(h(x)-g(x))^2]$. This bound is in the same spirit of \cite[Theorem 4.3]{staib2017distributionally}, which was derived for the original DRO problem (without domain adaptation).
\begin{theorem}
Let the labeling function $g$ satisfy $\|g\|_\sigma \leq \eta$. Therefore, for any $\delta >0$, with probability $1-\delta$, the following holds for all functions $h$ satisfying that $\|h\|_\sigma \leq \eta$:
\begin{equation}\label{eqn: G_bound}
    R_t \leq B\hat{R}_s +4\eta^2 \left(\sqrt{\frac{B^2}{n_s}+\frac{1}{n_t}}+\sqrt{\frac{1}{n_t}}\right ) \left(1+\sqrt{2\log\frac{1}{\delta}} \right)
\end{equation}
\proof 
We denote the empirical weighted source domain risk by $\hat{R}_{t'}:=\mathbb{E}^ {\mathbf{\hat{P}_t'}}[(h(x)-g(x))^2]$.  Based on the choice of the radius $\epsilon_{t'}$ in \eqref{eqn: main results}, for any $\delta > 0$, with probability $1-\delta$, we have that 
\begin{align}
    R_t&\leq  \hat{R}_{t'}+\epsilon_{t'}\|(h-g)^2\|_\sigma \nonumber\\
    &\leq \hat{R}_{t'} +\epsilon_{t'}(\|h^2\|_\frac{\sigma}{\sqrt{2}}+\|g^2\|_\frac{\sigma}{\sqrt{2}} + 2 \|h\|_\sigma\|g\|_\sigma)\nonumber\\ 
   &\leq \hat{R}_{t'} +4\eta^2 \epsilon_{t'}. 
\label{eq:bound_rt}
\end{align}
The first inequality follows from the first part of Theorem \ref{Thm} and \eqref{eqn: eq1}, and the second follows from the triangular inequality. The last inequality in \eqref{eq:bound_rt} is because $ \| h g\|_{\frac{\sigma}{\sqrt{2}} } \leq \| h\|_\sigma \| g\|_\sigma\leq \eta^2$ (Lemma~\ref{L3}), and hence  $\| h^2\|_{\frac{\sigma}{\sqrt{2}} },  \| g^2\|_{\frac{\sigma}{\sqrt{2}} }\leq \eta^2$. Since $\hat{R}_{t'}=\frac{1}{n_s}\sum_{i=1}^{n_s} \hat{w}(x_i) (h(x_i)-g(x_i))^2\leq\frac{1}{n_s}\sum_{i=1}^{n_s} B (h(x_i)-g(x_i))^2=B\hat{R}_s$, we can write 
\begin{equation}
    R_t\leq  B\hat{R}_{s} +4\eta^2 \epsilon_{t'}.
\end{equation}
Finally, since we are using a Gaussian kernel, we have $M=1$ in \eqref{eqn: main results}, which completes the proof.  
\end{theorem}

We note that the RHS of  \eqref{eqn: G_bound} is inversely proportional to the source and target domain samples sizes $n_s$ and $n_t$, and directly proportional to $B$. Therefore, for large sample sizes (i.e., $n_s, n_t\to\infty$), we have that $R_t\leq  B\hat{R}_{s}$, i.e., depends on $B$, which is indicative of the degree of discrepancy between both domains.  
\section{Experimental Results}
In this section, we verify the performance of the proposed approach.
 %
First, we generate data following the regression model $y=g(x)+n$, where $g(x)=k_\sigma(x,1)-k_\sigma(x,-1)$, $n$ is drawn from a normal distribution with zero mean and $0.1^2$ variance, and the source and target domain samples follow the distributions $\mathcal{N}(1,0.5^2)$ and $\mathcal{N}(-1,0.6^2)$, respectively. 
 
\begin{figure}[h] 
\centering
    \begin{subfigure}{0.45\textwidth}
    \includegraphics[width=\textwidth]{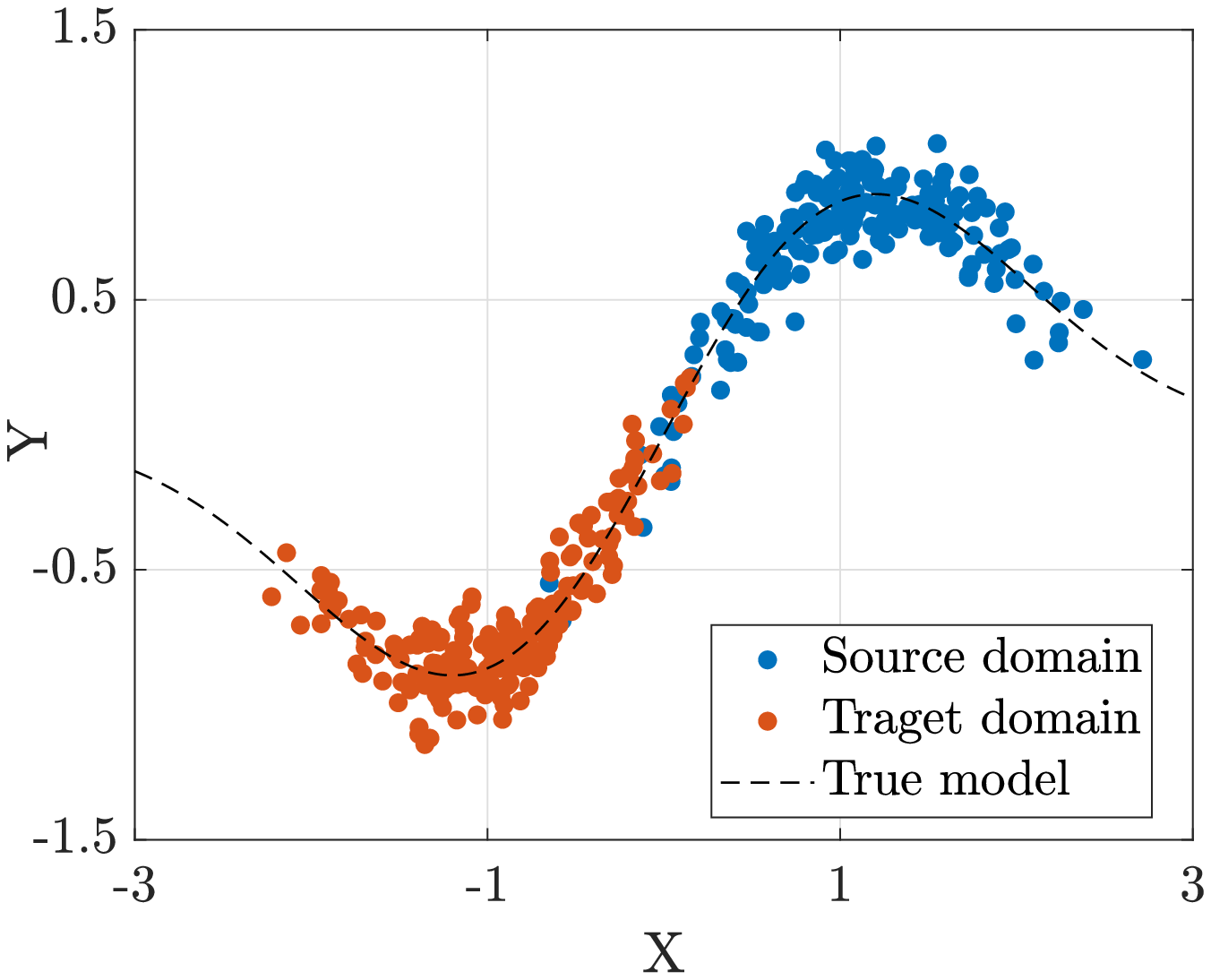}
    \caption{}
    \label{fig:Toy}
    \end{subfigure}
    \begin{subfigure}{0.45\textwidth}
    \centering
    \includegraphics[width=\textwidth] {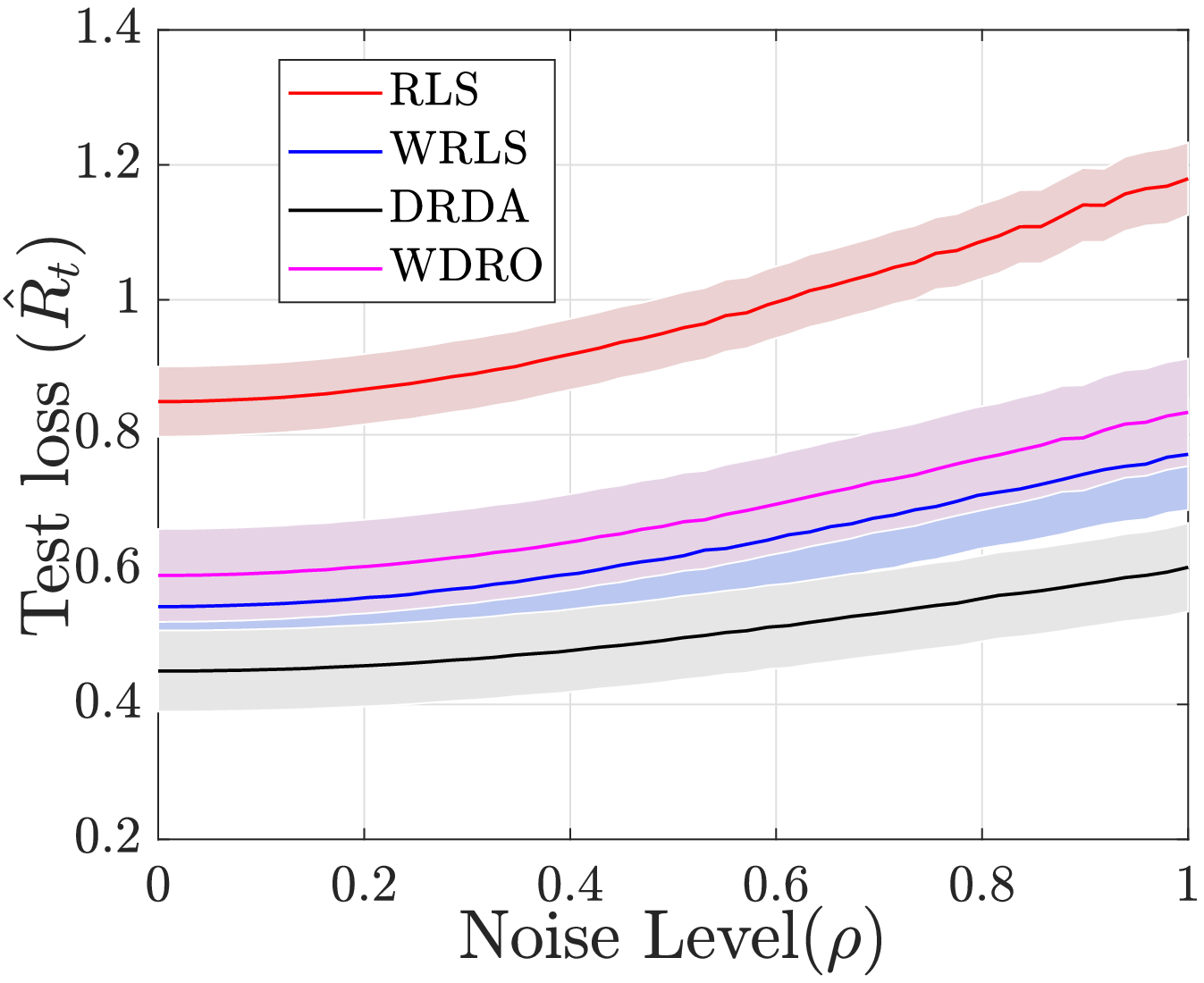}
    \caption{}
    \label{fig:exp1}
    \end{subfigure}
    \vspace{-0.2 cm}
    \caption{\small{(a) True model (dashed line), source (blue) and target (red) domain samples; (b) Test loss of different regression models as function of the noise level $\rho$.}}
    \label{fig:my_label}
    \vskip -0.1in
\end{figure}

 In our first experiment, we verify the robustness of our learned model to perturbations.  We sample $50$ source and target samples of size $100$. For each instance, we learn a regression model and test it on an unseen target domain sample $X_t$ of size $500$ for different noise levels. A depiction of these samples along with the true model are shown in Figure \ref{fig:Toy}.\par 
 
 We perturb $X_t$ with additive noise $\Delta\sim \mathcal{N}(0,\rho^2)$ with different noise levels $\rho$ , i.e., $\hat{X}_{t} = X_t + \Delta $. We compute the test loss (risk) $\hat{R}_{t}=\frac{1}{n_t}\sum_{i=1}^{n_{t}}l(h(x_i))$ for $\hat{X}_{t}$ for each noise level $\rho\in[0,1]$ and report their average loss $\hat{\mathbf{R}}_{t}$ and the corresponding $95\%$ interval. 
 We compare the performance of DRDA to different least-square regression approaches (see Table \ref{tab:LS formulationsl}). In the approach that we call weighted-DRO (W-DRO), we first solve for the weights $\hat{w}$ using \eqref{eqn: kmm}, then optimize over the decision function $h$ in \eqref{eqn: W DRO}, in contrast to the joint optimization in DRDA.  The hyperparameters $\beta$ and $\lambda$ are set to $10$ and $1.2$, respectively.

 Figure \ref{fig:exp1} demonstrates the test loss of the DRDA learned model for various noise levels in comparison to the least-square models. As shown, our DRDA model achieves the lowest average loss for all noise levels due to the built-in robust domain adaptation capability 
 along with the joint optimization over the weights and decision function in \eqref{eqn: DRDA}. To highlight the importance of the domain adaptation inherent in our framework, we also tested the standard DRO scheme, 
 which only uses the source sample for training the model. We found DRO without domain adaptation to be considerably less robust than DRDA in this setting. 
 For example, at $\rho=0.8$, the DRO model 
 yields an average test loss of $1.540$ versus $0.556$ for the DRDA model. Moreover, W-RLS and W-DRO, which first learn the weights then optimize over the model, underperform the DRDA model, underscoring the gain of jointly optimizing over $\hat{w}$ and $h$. 
   
\begin{table}[]
    \centering
    \scalebox{1}{\begin{tabular}{||c|c||} \hline\hline
         Method & Formulation  \\ \hline
         
          Regularized Least Squares (RLS) & $\min_{\alpha} \|\mathbf{K}^s_\sigma\alpha -y\|^2_2 + \lambda \alpha^T \mathbf{K}^s_\sigma \alpha$\\
          Weighted Regularized Least Squares (W-RLS)  &$ \min _{\alpha}(\mathbf{K}^s_\sigma\alpha-y)^T \mathbf{W} (\mathbf{K}^s_\sigma \alpha-y) + \lambda \alpha^T \mathbf{K}^s_\sigma \alpha$ \\
          Weighted DRO (W-DRO) & \eqref{eqn: W DRO}     \\
          DRDA  & \eqref{eqn: DRDA1} \\ \hline
    \end{tabular}}
    \vspace{0.1cm}
    \caption{\small{Different least-square methods ($\mathbf{W}$ in W-RLS and W-DRO is estimated using the KMM formulation in \eqref{eqn: kmm})}.}
    \label{tab:LS formulationsl}
    \vskip -0.15in
\end{table}

\begin{figure}[h] 
\centering
    \begin{subfigure}{0.45\textwidth}
    \includegraphics[width=\textwidth]{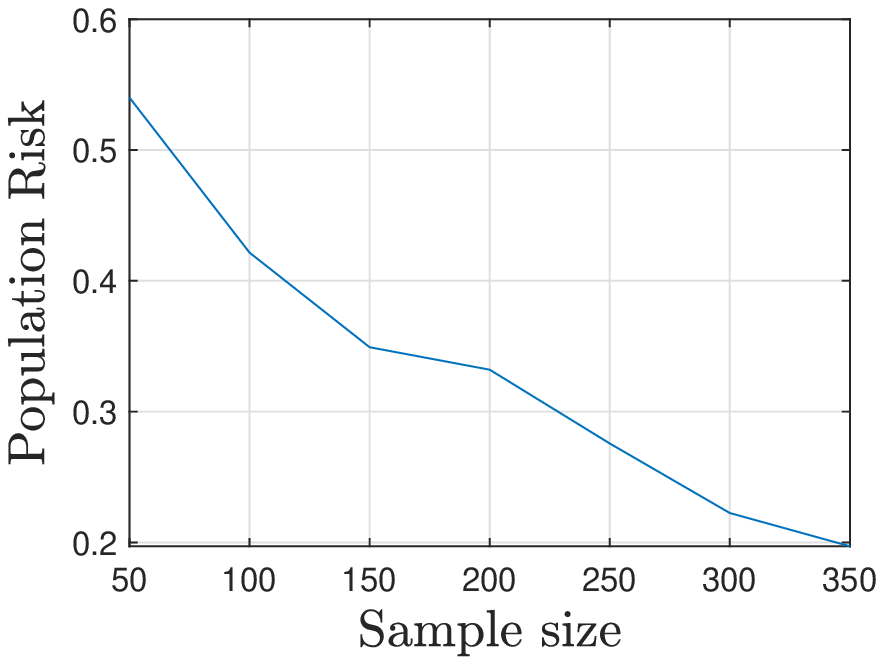}
    \caption{}
    \label{fig: T_Risk}
    \end{subfigure}
    \begin{subfigure}{0.45\textwidth}
    \centering
    \includegraphics[width=\textwidth] {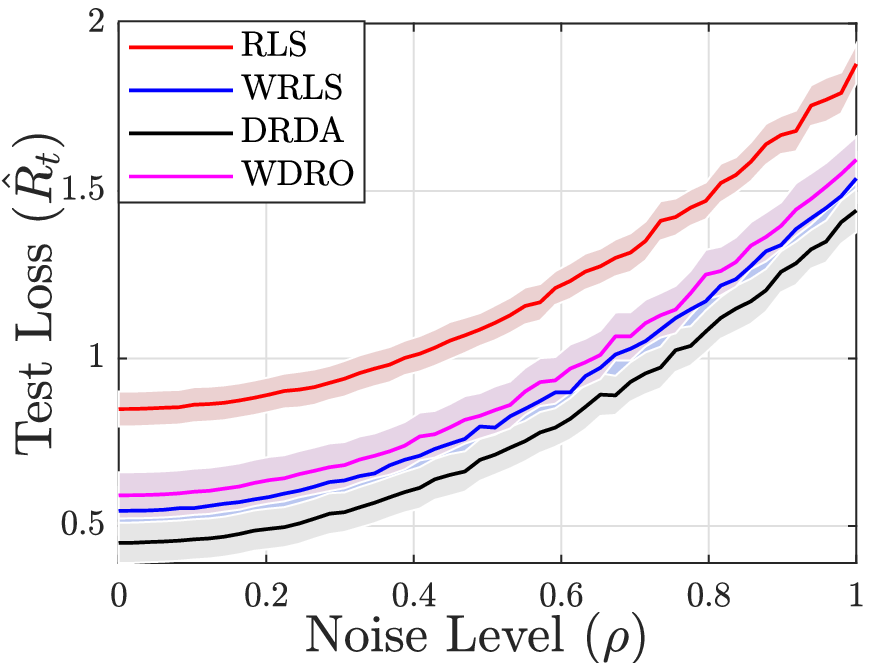}
    \caption{}
    \label{fig:exp3}
    \end{subfigure}
    \vspace{-0.2 cm}
    \caption{\small{(a) The true (population) target domain risk as a function of the sample size; (b) Test loss of different regression models as function of the noise level $\rho$}.}
    \label{fig:my_label}
    \vskip -0.1in
\end{figure}

 For the second experiment, we demonstrate the effect of the sample size on the target domain population risk. 
 We sample source and target domain samples of different sizes. For each sample size, we use the source and target domain samples to learn the DRDA model. Figure \ref{fig: T_Risk} shows the target domain risk as a function of the sample size. As expected, the risk decreases with the sample size, since training with a larger sample size (source and target) results in a model of higher accuracy.  \par

In our third experiment, we evaluate the performance of the proposed DRDA when the perturbations are added to the response. Specifically, $Y_t$ is perturbed with additive noise $\Delta$, i.e., $\hat{Y}_{t} = Y_t + \Delta $. As shown in Figure \ref{fig:exp3}, Our approach outperforms other regression approaches as it achieves the lowest average loss at all noise levels.
\section{Conclusion and Future Work}
Existing approaches to domain adaptation often fall short of yielding a decision model that is robust to perturbations and generalizes well to unseen target domain data. To address this limitation, we formulated a robust domain adaptation problem, dubbed Distributionally Robust Domain Adaptation (DRDA), that leverages a  DRO framework. Our formulation simultaneously accounts for domain adaptation and the uncertainty in the target domain sample. Since the target domain labels are unavailable, we re-weight the source domain sample to minimize the discrepancy between the two domains. Also, we constructed an uncertainty set, centered at the empirical weighted source domain distribution, and prove that it contains the true target domain distribution with high probability. In turn, the worst-case risk gives a high probability upper bound on the true population risk, thereby providing a guarantee on the generalization of the learned model. 
Our experimental results demonstrate that the learned regression model outperforms existing least-square approaches both in terms of robustness to noise and generalization power.

Our future work will explore extensions of the robust DA framework to other instances of robust learning and inference with distributional shifts, including classification and multi-output regression. Another avenue of future investigation will focus on relaxing the covariate shift assumption to account for shifts in the conditional distributions of the labels given the features in both domains. Therein, the key challenge will be to construct appropriate uncertainty sets to guarantee the robustness of the learned models and their out-of-sample performance, and to obtain bounds on the prediction and estimation errors of the solution. 
\section*{Funding}
NSF Award CCF-2106339; NSF CAREER Award CCF-1552497; DOE Award DE-EE0009152.

\section*{References}
\bibliography{ref.bib}
\bibliographystyle{plain}

\end{document}